\documentclass[11pt, a4paper, logo, copyright]{googledeepmind}

\usepackage[style=authoryear, backend=biber, natbib=true, maxcitenames=2, dashed=false]{biblatex}
\usepackage[capitalise]{cleveref}

\pdftrailerid{redacted}

\usepackage{kantlipsum, lipsum}
\usepackage{dsfont}
\usepackage{enumitem}

\usepackage{wrapfig}
\usepackage{gdm-colors}
\usepackage{subcaption}  
\usepackage[export]{adjustbox}  
\usepackage[algo2e, ruled,vlined]{algorithm2e}

\graphicspath{{figures/}}

\title{How Transparent is DiffusionGemma?}

\correspondingauthor{joshengels@google.com, mcdougallc@google.com}


\reportnumber{001} 

\renewcommand{\today}{2026-06}

\author{%
  Joshua Engels\textsuperscript{$\dagger$}, 
  Callum McDougall\textsuperscript{$\dagger$}, 
  Bilal Chughtai\textsuperscript{$\dagger$} 
  
  Janos Kramar, Senthooran Rajamanoharan, Cindy Wu, Arthur Conmy, Asic Q Chen, Jean Tarbouriech, Min Ma
  
  Brendan O'Donoghue\textsuperscript{$\ddagger$},
  João Gabriel Lopes de Oliveira\textsuperscript{$\ddagger$},  
  Rohin Shah\textsuperscript{$\ddagger$}, 
  Neel Nanda\textsuperscript{$\ddagger$} \vspace{0.2cm}
}

\affil[\ ]{Google DeepMind}
\affil[$\dagger$]{Primary Contributor}
\affil[$\ddagger$]{Advising}

\usepackage{mathtools}
\usepackage{dsfont}
\usepackage[dvipsnames]{xcolor}
\usepackage[colorinlistoftodos]{todonotes}
\usepackage{booktabs}
\usepackage{xfrac}
\usepackage{bbm}

\usepackage{algorithm}
\usepackage{algorithmicx}

\usepackage{xparse}
\usepackage{lipsum}
\usepackage{changepage}
\usepackage{enumitem}

\newcommand{\squishlist}{
   \begin{list}{$\bullet$}
    { \setlength{\itemsep}{0pt}      \setlength{\parsep}{3pt}
      \setlength{\topsep}{3pt}       \setlength{\partopsep}{0pt}
      \setlength{\leftmargin}{1.5em} \setlength{\labelwidth}{1em}
      \setlength{\labelsep}{0.5em} } }

\newcommand{\squishlisttwo}{
   \begin{list}{$\bullet$}
    { \setlength{\itemsep}{0pt}    \setlength{\parsep}{0pt}
      \setlength{\topsep}{0pt}     \setlength{\partopsep}{0pt}
      \setlength{\leftmargin}{2em} \setlength{\labelwidth}{1.5em}
      \setlength{\labelsep}{0.5em} } }

\newcommand{\squishend}{
    \end{list}  }

\newcommand{\myvec}[1]{\boldsymbol{#1}}

\newcommand{\two}{(2)}

\newcommand{\vo}{\myvec{o}}

\newcommand{\vs}{\myvec{s}}

\newcommand{\vQ}{\myvec{Q}}

\newcommand{\vS}{\myvec{S}}

\newcommand{\vW}{\myvec{W}}

\newcommand{\vZ}{\myvec{Z}}

\DeclareMathAlphabet{\mathpzc}{OT1}{pzc}{m}{n}

\usepackage{xcolor}

\usepackage{multirow}
\usepackage{float}  
\usepackage{longtable}  
\usepackage{xurl}

\begin{document}

\begin{abstract}
LLM reasoning transparency is a critical affordance for understanding model decisions, mitigating misuse and misalignment, and debugging surprising model behaviors. However, DiffusionGemma performs a larger fraction of its computation in a continuous latent space; does this make its reasoning less transparent? We study this question by decomposing transparency into two components: \textit{variable transparency}, whether we understand intermediate snapshots of a model's computational state; and \textit{algorithmic transparency}, whether we can use these snapshots to reconstruct the process by which the model arrived at its outputs. Naively, DiffusionGemma has poor variable transparency: its opaque serial depth, the amount of serial computation that occurs in between interpretable model states, seems at first 28.6X higher than the corresponding autoregressive Gemma 4 model. However, we show that we can map the information flowing between denoising steps through an interpretable token bottleneck with no decrease in downstream performance. Treating these intermediate states as interpretable reduces the opaque serial depth to just 1.1X that of Gemma 4. Algorithmic transparency is harder for diffusion models than for autoregressive models because all token predictions in the canvas can change at every denoising step, giving the model the power to implement complicated distributed algorithms during the denoising process. To begin bridging this gap, we conduct a suite of interpretability case studies, uncovering initial evidence of novel diffusion-specific phenomena such as non-chronological reasoning, token and sequence smearing, and intermediate-context reasoning. Finally, we test \textit{monitorability}, a key application of transparency that measures whether model outputs are useful for downstream tasks. We find that DiffusionGemma is similarly monitorable to Gemma 4. 
\end{abstract}

\maketitle

\section{Introduction}

Most current frontier AI systems are autoregressive reasoning models that reason in natural language via \textit{chain of thought} (CoT). Maintaining human-understandable chain of thought is important for many reasons: for example, model misbehavior like reward hacking \citep{baker2025monitoring} and prompt injections \citep{chennabasappa2025llamafirewall} are frequently only apparent in CoT; monitors with access to CoT strongly outperform monitors without CoT on monitorability evaluations \citep{guan2025monitoring, arnav2025cot}; and resampling CoT provides a way to understand the structure of a model's reasoning process \citep{bogdan2025thoughtanchorsllmreasoning}. Chain of thought transparency is particularly important for AI safety: we can give the CoT to another model and task it with detecting signs of misuse and misalignment \citep{baker2025monitoring, guan2025monitoring, korbak2025chain, arnav2025cot}. Indeed, CoT transparency may in the future become load-bearing in determining whether capable AI systems are safe to deploy; for example, existing proposals for AI control safety cases rely heavily on being able to effectively monitor deployments \citep{korbak2025sketch, metr-2026-frontier-risk-report, anthropic2026risk, emmons2025chain}.

DiffusionGemma \citep{diffusiongemmamodelcard2026} is a newly released \textit{text diffusion model}. Text diffusion models iteratively refine a large \textit{canvas} of many tokens by repeatedly applying a denoising operation to the entire canvas. Importantly, DiffusionGemma's intermediate states between denoising steps are not purely natural language: they consist partially of dense \textit{self-conditioning} vectors that are not by default human-interpretable. 

This means that DiffusionGemma may not ``think in English'': for example, it is possible for DiffusionGemma to consider multiple options at once for a given token without this being visible in its chain of thought. Additionally, because DiffusionGemma is not autoregressive, it is possible for it to ``change its mind'' on a token during denoising, and for later tokens in the canvas to causally influence earlier tokens in the canvas, again without any of this causal process being reflected in natural language. We thus might expect DiffusionGemma, and future models like it, to be less transparent than a standard autoregressive model.

However, the intermediate outputs of each denoising step might, with some work, allow us to recover a similar level of transparency as autoregressive models. In this work, we study the question of how transparent DiffusionGemma is and whether we can improve its transparency by utilizing this inter-step information.

\begin{figure}[!t]
    \centering
    \includegraphics[width=\linewidth]{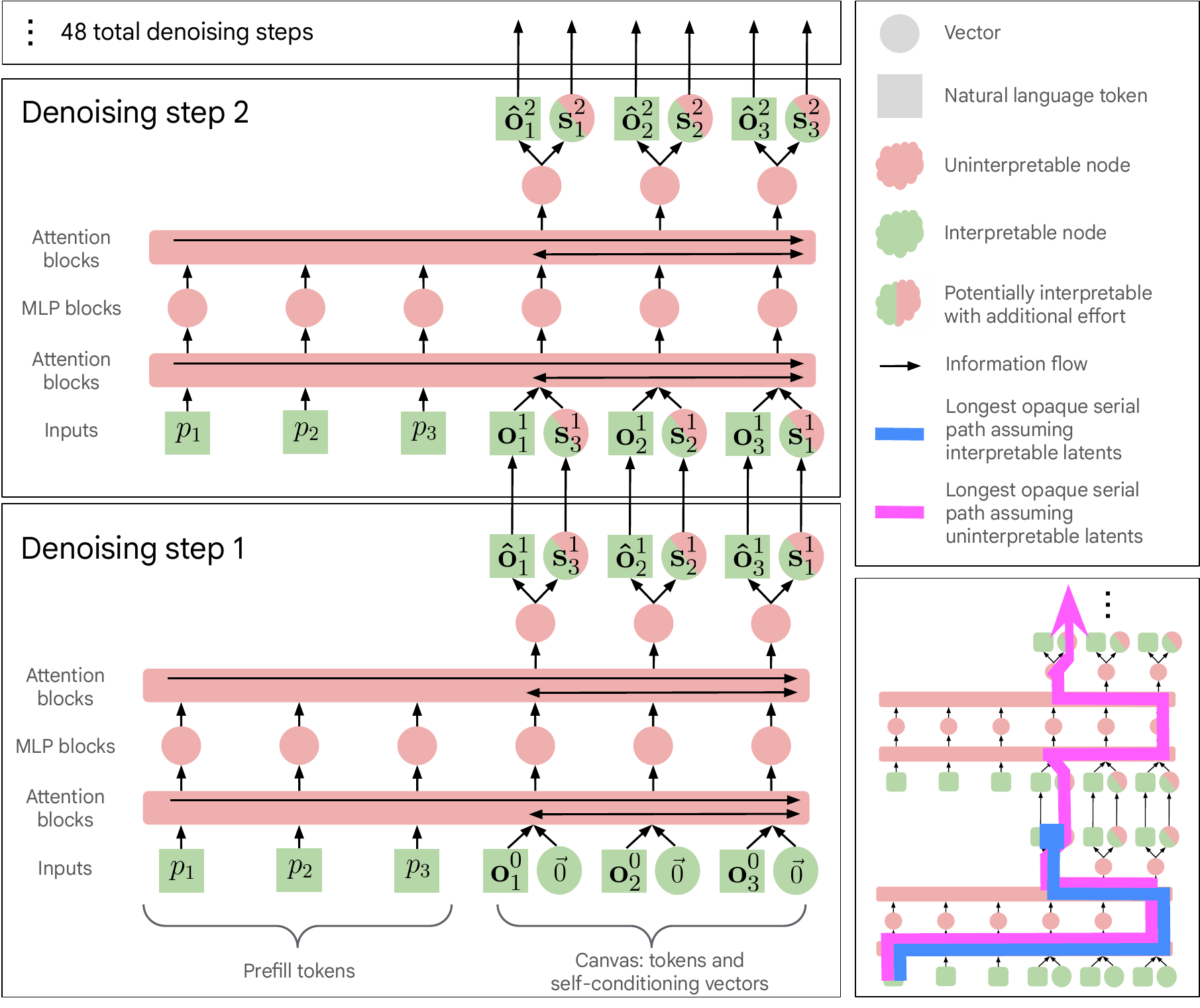}
    \caption{A simplified architecture diagram of DiffusionGemma, with the first two denoising steps shown; see \Cref{sec:architecture} for an in-depth description of DiffusionGemma's architecture and sampling. We additionally show the path with the largest \textit{opaque serial depth} with and without the assumption that the intermediates $\vs$ are interpretable; see \Cref{sec:depth_calc} for more details on this calculation.}
    \label{fig:main}
\end{figure}

We break down transparency along the following axes:
\begin{enumerate}[align=left,leftmargin=*]

    \item \textbf{Opaque Serial Depth:} In \Cref{sec:serial_depth}, we review the architecture of DiffusionGemma and study its \textit{opaque serial depth}: the length of the longest serial computation that is performed in the architecture without passing through an interpretable bottleneck \citep{brown2026quantifying}. DiffusionGemma has a bottleneck between denoising steps comprising $C$ natural language tokens and $C$ self-conditioning vectors, where $C$ is the size of the canvas. We show that without an assumption that this bottleneck is interpretable, DiffusionGemma carries out $28.6$X more opaque serial computation than the corresponding Gemma model. However, if indeed these intermediate steps are interpretable, the opaque serial depth drops to $1.1$X that of the corresponding Gemma model. 
    
    \item \textbf{Variable Transparency:} \Cref{sec:serial_depth} establishes the serial depth of DiffusionGemma \textit{if} the bottlenecks are interpretable; we thus next focus on determining \textit{whether} they are. In \Cref{sec:bottleneck_ablation}, we show that we can reduce the information flow in the bottlenecks between denoising steps to $O(c)$ natural language tokens with minimal performance degradation. In \Cref{sec:top_tokens}, we further show that these $O(c)$ tokens are usually guesses for nearby token positions, implying that these mapped tokens are frequently interpretable.
    
    \item \textbf{Monitorability:} \cite{guan2025monitoring} introduce a suite of monitorability evaluations that study the informativeness of a model's chain of thought via the proxy task of monitors' ability to extract said information. In \Cref{sec:monitorability} we adapt these evaluations and apply them to DiffusionGemma, finding that DiffusionGemma and Gemma 4 are similarly monitorable. 
    
    \item \textbf{Algorithmic Transparency:} DiffusionGemma has a substantially different architecture\footnote{Here and elsewhere in the paper we use the word ``architecture'' to refer to the entire inference-time computational graph of a model, e.g. the base transformer architecture (which is the same for Gemma and DiffusionGemma) \textit{and} the repeated denoising forward steps, the parameters to project the self conditioning vectors, etc.} to standard autoregressive transformers. As such, it is capable of reasoning in ways autoregressive models cannot. In \Cref{sec:case_studies}, we present a series of case studies that highlight the most interesting differences that we found. We think scientific understanding of these differences could be important in understanding and predicting future behaviors of similar, potentially more powerful models. We explore a number of interesting phenomena unique to text diffusion models, including non-chronological reasoning, token and sequence smearing, and intermediate-context reasoning.
\end{enumerate}

Finally, we note that while we believe that DiffusionGemma retains most of the transparency benefits of autoregressive CoT, this need not apply in future latent reasoning models. Our results rely heavily on DiffusionGemma's specific architectural and training choices; for example, DiffusionGemma has a projection in intermediate layers that biases the vector that gets passed between denoising steps towards the embedding vectors. If future models are trained differently or have different architectures, or if we did for example a large quantity of outcome-based RL, we might find that models' latent spaces become significantly more opaque (see Appendix A.2 of \citet{brown2026quantifying} for more discussion of the interaction of training techniques and human-understandable information). We hope that model developers use the methods explored in this paper to continuously evaluate future models and detect regressions in transparency and monitorability.

\section{Opaque Serial Depth}
\label{sec:serial_depth}

A fundamental worry about latent reasoning architectures is that they may enable models to perform more non-transparent reasoning in latent space than current architectures. The amount of serial computation a model can perform without passing through an interpretable text bottleneck can be formalized via its \textit{opaque serial depth} \citep{browncohen2026quantifyingnecessitychainthought}. Consider the Boolean circuit that computes the same function as a neural network (including its sampling procedure) and then mark the nodes in the circuit that we can interpret; a model's opaque serial depth is then loosely defined as the longest path in this circuit that does not contain any interpretable nodes. See \citet{browncohen2026quantifyingnecessitychainthought} for a more rigorous discussion. 

The opaque serial depth of a model is a function of its architecture, so we begin this section by describing the architecture of DiffusionGemma. We will then compute upper bounds on the opaque serial depth of DiffusionGemma and Gemma both empirically and asymptotically.

\subsection{DiffusionGemma Architecture}
\label{sec:architecture}

DiffusionGemma adapts the architecture of the Gemma 4 26B A4B model~\citep{gemma4modelcard2026}. We discuss the main differences briefly in this section and include a more detailed discussion in \Cref{app:arch}. The architecture is also visualized in \Cref{fig:main}.

DiffusionGemma inference consists of a \textbf{setup} stage followed by $T$ iterations of a \textbf{denoising loop}, each step consisting of a \textbf{forward pass} and a \textbf{sampling step}.

Unlike the Gemma family of models, which generate a single token at a time, DiffusionGemma takes as input a user prompt $\mathbf{p} = (p_1, \ldots, p_N) \in [V]^{N}$ and produces an output \textit{canvas}, consisting of $C$ tokens $\mathbf{o}^T = (\vo_1^T, \ldots, \vo_C^T) \in [V]^{C}$. Between each iteration of the denoising loop, we pass the current canvas $\mathbf{o}^t \in [V]^{C}$ and a \textit{self-conditioning matrix} $\vS^{t} \in \mathbb{R}^{C \times d_\text{model}}$ to the next denoising step. These two quantities form an information \textit{bottleneck} between denoising steps.

\textbf{Setup:} The prompt is passed through the model once to get key-value activations; these activations are static for the denoising loop and attended back to during each sampling step's forward pass. The canvas is initialized to $C$ tokens sampled uniformly at random and $\vS^0 \in \mathbb{R}^{C \times d_\text{model}}$ is initialized to the zero matrix.

\textbf{Denoising Loop:} The model then iterates for $T$ denoising steps, indexed by $t \in \{1, \ldots, T\}$, where $t = 1$ is the first denoising step and $t = T$ is the last denoising step. In each denoising step $t$, the model takes as input the canvas $\vo^{t-1} \in [V]^{C}$ and the self-conditioning matrix $\mathbf{S}^{t-1} \in \mathbb{R}^{C \times d_\text{model}}$ from the previous step and produces updated canvas tokens $\vo^t \in [V]^{C}$ and an updated self-conditioning matrix $\mathbf{S}^t \in \mathbb{R}^{C \times d_\text{model}}$. Unless stated otherwise, we use adaptive stopping, which stops sampling when the entropy in predicted tokens is low and model predictions are no longer changing.

\textbf{Denoising Loop; Forward Pass:} We pass $\mathbf{S}^{t-1}$ through a gated MLP and then add it to the embeddings of the canvas tokens $\vo^{t-1}$. We then proceed through the rest of the transformer (which notably uses bidirectional attention) until we reach the final predicted logits $\boldsymbol{\ell}^t \in \mathbb{R}^{C \times |V|}$.

\textbf{Denoising Loop; Sampling Step:} We first compute the temperature $\tau_t$ to use for the step, which is a linear interpolation (keyed on the denoising step) between $\tau_{\max} = 0.8$ and $\tau_{\min} = 0.4$. We apply $\tau_t$ to the logits to get $\hat{\boldsymbol{\ell}}^t = \boldsymbol{\ell}^t / \tau_t$. 

The new self-conditioning matrix $\mathbf{S}^t$ is computed by embedding the shaped logits back into the model's embedding space via a soft embedding:
\begin{equation}
    \label{eqn:softmax}
  \mathbf{S}^t = \text{softmax}(\hat{\boldsymbol{\ell}}^t) \cdot \mathbf{W}_E \in \mathbb{R}^{C \times d_\text{model}}
\end{equation}

The self-conditioning matrix is thus a transformed version of the last activation from the previous denoising step, but weighted towards the directions in activation space corresponding to more likely tokens. 

Candidate tokens are first sampled independently at each position from the shaped logits: 
\begin{align}
    \hat{\vo}_i^t \sim \text{Categorical}(\text{softmax}(\hat{\boldsymbol{\ell}}^t_i)) \text{ for } i \in \{1, \ldots, C\}
\end{align}

We then compute $\vo^t$ via Entropy-Bounded (EB) sampling~\citep{ben2026accelerated}, which keeps low entropy tokens and renoises high entropy tokens; see \Cref{app:arch} for more details.

\textbf{Final output:} After $T$ steps, the final canvas tokens are obtained by taking the argmax of the logits: $\vo_i^T = \arg\max_v \boldsymbol{\ell}^T_{i,v}$ for $i \in \{1, \ldots, C\}$.

\textbf{Multi-canvas sampling:} Since the canvas has a fixed length $C$, generating outputs longer than $C$ tokens requires multi-canvas sampling. After completing the denoising loop for one canvas, the output tokens $o^T$ are appended to the original prompt tokens $\mathbf{p}$ to form a new prompt $\mathbf{p}' = (p_1, \ldots, p_N, o_1^T, \ldots, o_C^T) \in [V]^{N+C}$, and a fresh canvas is then sampled.

\subsection{Opaque Serial Depth Calculation}
\label{sec:depth_calc}

We now compute the opaque serial depth of Gemma 4 models and DiffusionGemma. We compute the exact asymptotic opaque serial depths by hand (see \Cref{app:asymptotic} for details) and compute an empirical upper bound using an open source serial depth analyzer\footnote{Available at \href{https://github.com/google-deepmind/serial_depth}{https://github.com/google-deepmind/serial\_depth}}.

When computing the opaque serial depth, we make the assumption that the natural language inputs and outputs of the model are interpretable because DiffusionGemma chains of thought and responses are qualitatively similar to autoregressive models and are unlikely to be obfuscated. We then need to decide about whether we can assume intermediate tokens $o^t_i \in [V]^{C}$ and self conditioning vectors $\vs^t \in \mathbb{R}^{C\times d_{model}}$ are interpretable. We will come back to this question in later sections of the paper, but for now, we compute the opaque serial depth in both cases.

We compute the opaque serial depth at a sequence length of 256k for all models. We consider DiffusionGemma at $T=48$ steps for the uninterpretable bottleneck case and at $T=1$ step for the interpretable bottleneck case. Our results are shown in \Cref{tab:serial-depth-profiling}. Note that setting $T = 48$ is the worst case for DiffusionGemma; in practice adaptive stopping means that we use only $12$ to $16$ steps, so the usual opaque serial depth may be about a third as large as reported in the table.
\begin{table}[t]
    \centering
    \caption{Empirical upper bounds and asymptotic computations of the opaque serial depths of Gemma 4 and DiffusionGemma evaluated on 256k context sequences. For the asymptotic calculation, $T$ is the number of denoising steps, $L$ is the number of layers, $C$ is the size of the canvas, $N$ is the number of context tokens, and $D$ is the dimension of activations. UB stands for "Upper Bound".}
    \label{tab:serial-depth-profiling}
    \begin{tabular}{lcl}
        \toprule
        \textbf{Model} & \begin{tabular}{@{}l@{}}\textbf{Empirical Opaque} \\ \textbf{Serial Depth (UB)}\end{tabular} & \begin{tabular}{@{}l@{}}\textbf{Asymptotic Opaque} \\ \textbf{Serial Depth}\end{tabular} \\
        \midrule
        Gemma 4 E2B & 8,978 & $\mathcal{O}\left( L \cdot (\log N + \log D) \right)$ \\
        Gemma 4 E4B & 10,886 & $\mathcal{O}\left( L \cdot (\log N + \log D) \right)$ \\
        Gemma 4 26B A4B & 21,235 & $\mathcal{O}\left( L \cdot (\log N + \log D) \right)$ \\
        Gemma 4 31B & 13,750 & $\mathcal{O}\left( L \cdot (\log N + \log D) \right)$ \\
        \midrule
        \begin{tabular}{@{}l@{}}DiffusionGemma 26B A4B \\ Interpretable Bottleneck\end{tabular} & 23,571 & $\mathcal{O}\left( L \cdot (\log N + \log D + \log C ) \right)$ \\
        \begin{tabular}{@{}l@{}}DiffusionGemma 26B A4B\\ Uninterpretable Bottleneck\end{tabular} & 608,016  & $\mathcal{O}\left( T \cdot L \cdot (\log N + \log D + \log C )\right)$ \\
        \bottomrule
    \end{tabular}
\end{table}

Using the numbers in \cref{tab:serial-depth-profiling}, we find that the empirical opaque serial depth upper bound of DiffusionGemma with an uninterpretable bottleneck is $608,016$ $/$ $21,235$ $=28.6$X larger than Gemma 4 26B A4B, the corresponding Gemma model. Asymptotically, the opaque serial depth is approximately $T$ times larger. However, if we assume that the bottleneck is interpretable, the empirical opaque serial depth upper bound drops to $23,571$ $/$ $21,235$ $=1.1$X that of Gemma 4 26B A4B, and the asymptotic opaque serial depth is only greater by $\mathcal{O}(L \log C)$.

\section{Is DiffusionGemma's bottleneck interpretable?}

One of the key reasons for optimism for chain of thought monitorability is that the chain of thought is an interpretable intermediate node in the model's computation. In other words, for current autoregressive models, long serial chains of cognition must pass through the chain of thought, which is constructed out of interpretable tokens. Analogously, we might hope that the same is true of intermediate states in a text diffusion model. 

Recall that for DiffusionGemma, the information bottleneck between each denoising step consists of a canvas of tokens $\vo$ and a self-conditioning matrix $\vS^t \in \mathbb{R}^{C \times d_{model}}$. The tokens are similarly interpretable to normal autoregressive tokens--we can just read them, and as we will see at the end of this section these top tokens are usually identical or similar to the final predicted tokens--but $\vS^t$ is not. We wish to determine whether we can convert $\vS^t$ into interpretable tokens without a large loss in performance. 

We may hope that converting $\vS^t$ to interpretable tokens will be tractable because of \Cref{eqn:softmax}: multiplying by the embedding matrix and then applying the softmax biases information in $\vS^t$ to be aligned with the embedding directions, which correspond to interpretable tokens. We thus perform experiments that restrict the information in $\vS^t$ to just a few tokens per sequence position using the Logit Lens~\citep{nostalgebraist2020logitlens}. We compare the performance of this restricted version of DiffusionGemma to the unmodified model on a set of representative capability benchmarks: Natural2Code \citep{pichai2023gemini}, LiveCodeBench \citep{jain2024livecodebench}, AMC/AIME/IMO variants, and GPQA \citep{rein2024gpqa}. We enable dynamic thinking for all evals. At the end of this section, we run additional experiments that examine whether these tokens the Logit Lens projects out are indeed interpretable.

\begin{figure}[!t]
\centering
\includegraphics[width=\textwidth]{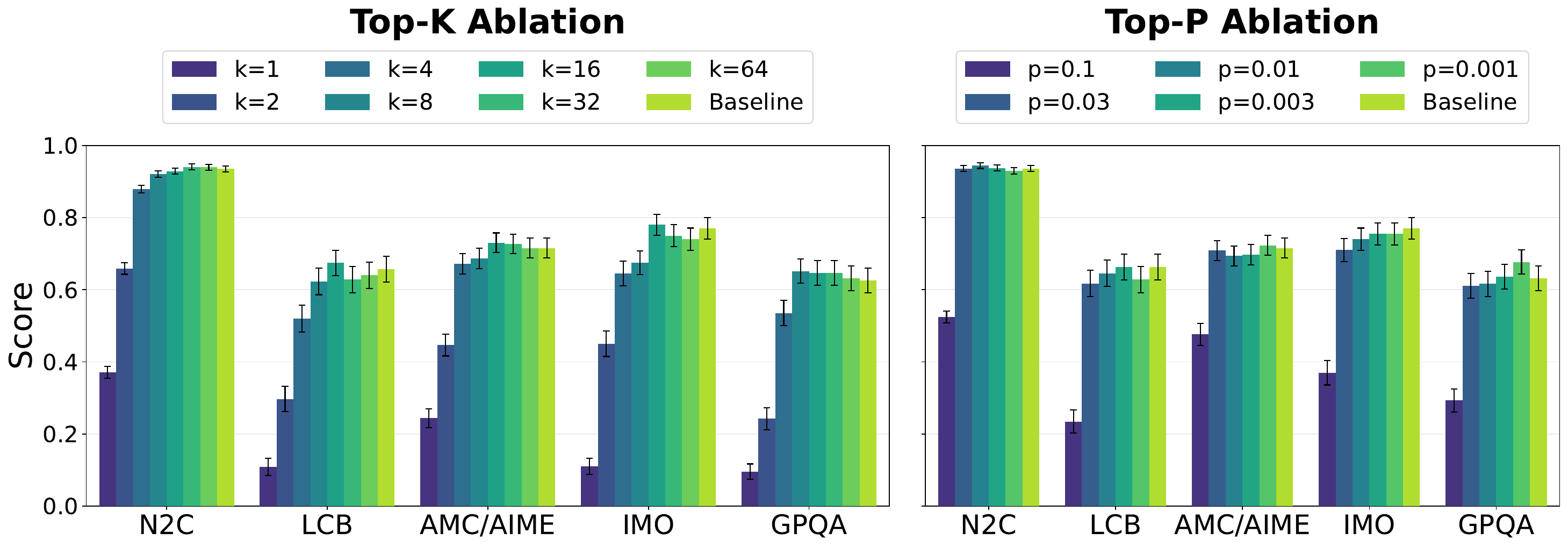}
\caption{We can restrict DiffusionGemma to use only the few most probable tokens at each denoising step without significantly harming performance on capability benchmarks.}
\label{fig:ablation_experiment}
\end{figure}

\subsection{Running Ablations on the Bottleneck}
\label{sec:bottleneck_ablation}
Recall from \cref{eqn:softmax} that the canvas condition matrix $\vS_{t}$ is computed as $\mathbf{S}^t = \text{softmax}(\hat{\boldsymbol{\ell}}^t) \cdot \mathbf{W}_E$. We now describe our method to restrict the information in $\vS^t$ to $k$ tokens. We consider logit modification functions $f: \mathbb{R}^{C \times |V|} \rightarrow \mathbb{R}^{C \times |V|}$, where we apply the function to get $\vS^{t'}$ as follows:
\begin{equation}
\vS^{t'} = \vW_E \cdot \texttt{Softmax} \left( f\left[\hat{\boldsymbol{\ell}}^t\right]\right)
\end{equation}
We experiment with different $f$. One such $f$ is $f_{p}$, which sets all logits that correspond to probabilities below $p$ to a constant value, such that after the softmax the probability of the retained tokens are the same as they were without the ablation (see \Cref{alg:f_p} in \Cref{app:bottleneck_appendix}). 
We also experiment with $f_{k}$, which sets all logits besides the top $k$ to a constant value in the same way as $f_p$. One downside of this technique is that a uniform distribution over low-probability logits is out of distribution for the model and so might overestimate the effect of the intervention; alternatives include setting other logits to $-\infty$, or setting them to be the unigram, bigram, or trigram token distribution. In our experiments, however, we always follow the model of \Cref{app:bottleneck_appendix} and set low probability logits such that the probabilities of the non-modified tokens stay the same.

\begin{figure}[t]
\centering
\includegraphics[width=\textwidth]{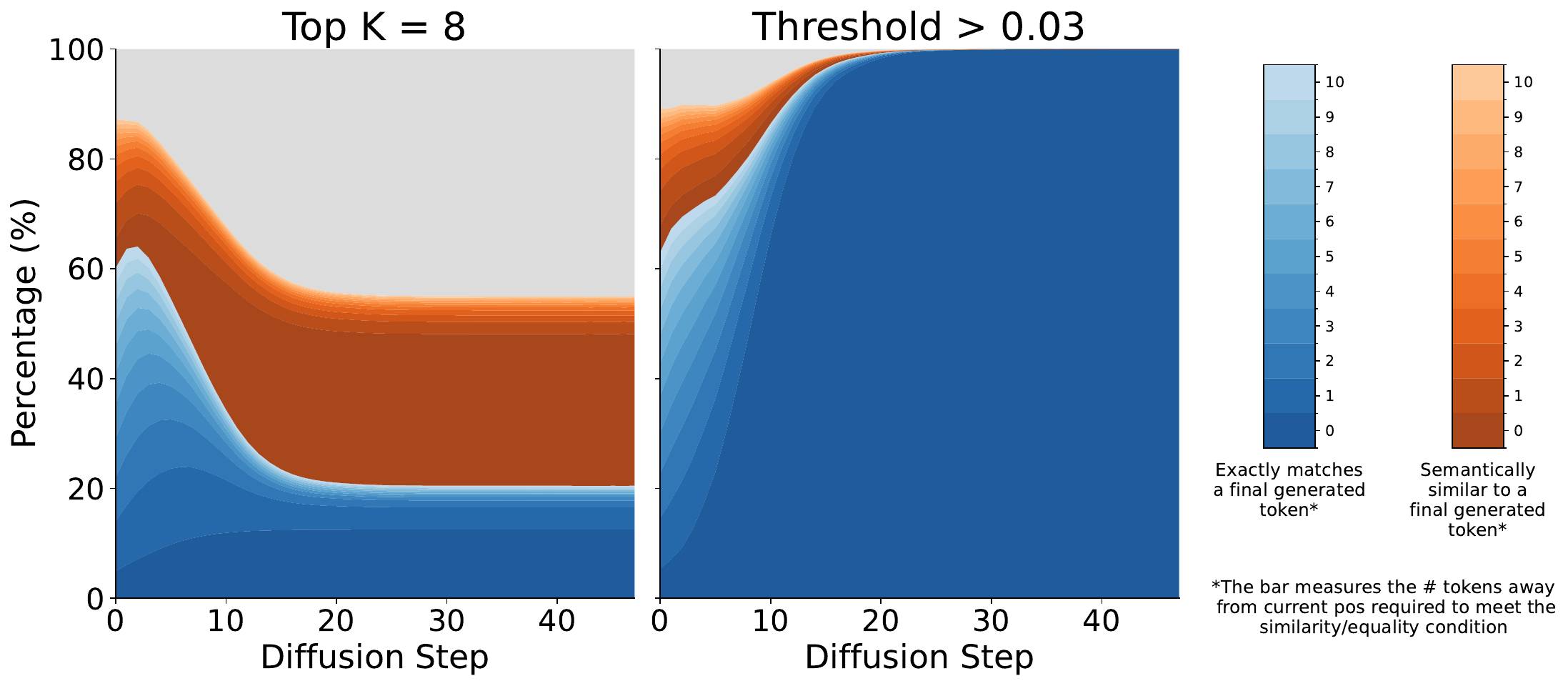}
\caption{Breakdown of intermediate state top token identities with different restrictions averaged across WildChat prompts. The set of tokens with probability $> 0.3$ are mostly either equal to or similar to the final token at the current position or adjacent positions. Top K = 8 has many uninterpretable tokens because the model is frequently highly confident in the top token at late denoising steps, and so the 7 non-top-1 tokens are more arbitrary.}
\label{fig:topk_token_investigation}
\end{figure}

In \Cref{fig:ablation_experiment} we report the results of capability evaluations using $f = f_{p}$ and $f = f_{k}$ to replace $\vS^t$ with $\vS^{t'}$ for all $t$ and all token positions. Increasing $k$ and decreasing $p$ tends to increase performance; $k = 8$ and $p = 0.03$ are enough to attain the same performance as the baseline.

If the projected tokens are interpretable under these interventions, then we have successfully reduced the serial depth by a factor of $26$. Thus, the next important question is whether these top-k tokens really are interpretable, which we now investigate.

\subsection{What Are the Top Tokens?}
\label{sec:top_tokens}
Early qualitative investigations showed that the top intermediate tokens were mostly guesses as to what the final tokens would be; that is, there was not that much inter-denoising step reasoning happening in the canvas. To quantify this intuition, in this section we test the proportion of top tokens that consist of these guesses. This also serves as a proxy measure for token interpretability; the more information that is simply passing guesses for token identities, the more confident we can be that \textit{most} information is reasonably interpretable.

We generate DiffusionGemma rollouts on $n=800$ user prompts from the WildChat dataset \citep{zhao2024wildchat} up to a maximum of $1024$ generated tokens (four canvases) with no early stopping and $T = 48$. At each denoising step $t$ and sequence position $i$, we take the results of the softmax in \Cref{eqn:softmax} and extract the set of tokens that meet one of the following conditions:

\begin{enumerate}[align=left,leftmargin=*]
    \item The token has a probability in the top 8 probabilities. The corresponding intervention is $f_{k=8}$.
    \item The token has a probability that is greater than 0.03. The corresponding intervention is $f_{p=0.03}$.
\end{enumerate}
These interventions correspond to the strictest interventions that do not harm capabilities in \cref{fig:ablation_experiment}.
We then assign each extracted token to one of the following categories (in decreasing order of priority):
\begin{enumerate}[align=left,leftmargin=*]
    \item \textbf{Final Token:} The token exactly matches the final token generated at the current position $i$.
    \item \textbf{Adjacent Token:} The token exactly matches a final token generated at a nearby position (up to $\pm 10$ positions). This captures instances where the model is considering a token but has not yet resolved its exact placement in the sequence.
    \item \textbf{Similar to Final:} The token is semantically similar to the final token at position $i$, where being semantically similar is defined as being among the 50 nearest neighbors by cosine-similarity of the target token in the Gemma 4 31B embedding matrix.
    \item \textbf{Similar to Adjacent:} The token is semantically similar to a final token at a nearby position.
    \item \textbf{Other:} The token does not fit into any of the above categories and is considered uninterpretable.
\end{enumerate}
We also ignore token positions that have an EOS. We plot category percentages versus denoising step (averaged across all token positions and prompts) in \Cref{fig:topk_token_investigation}. 

We find that the Top $K=8$ filter yields many uninterpretable tokens in the ``Other'' category, particularly at later denoising steps. As diffusion progresses, the model narrows its confidence to a single highly probable token, rendering the remaining 7 tokens in the top-8 set as low-probability noise. While these tokens may be noise, they remain a monitorability concern under this intervention because we cannot rule out that they convey opaque reasoning.

\begin{figure}[!t]
    \centering
    \includegraphics[width=0.75\linewidth]{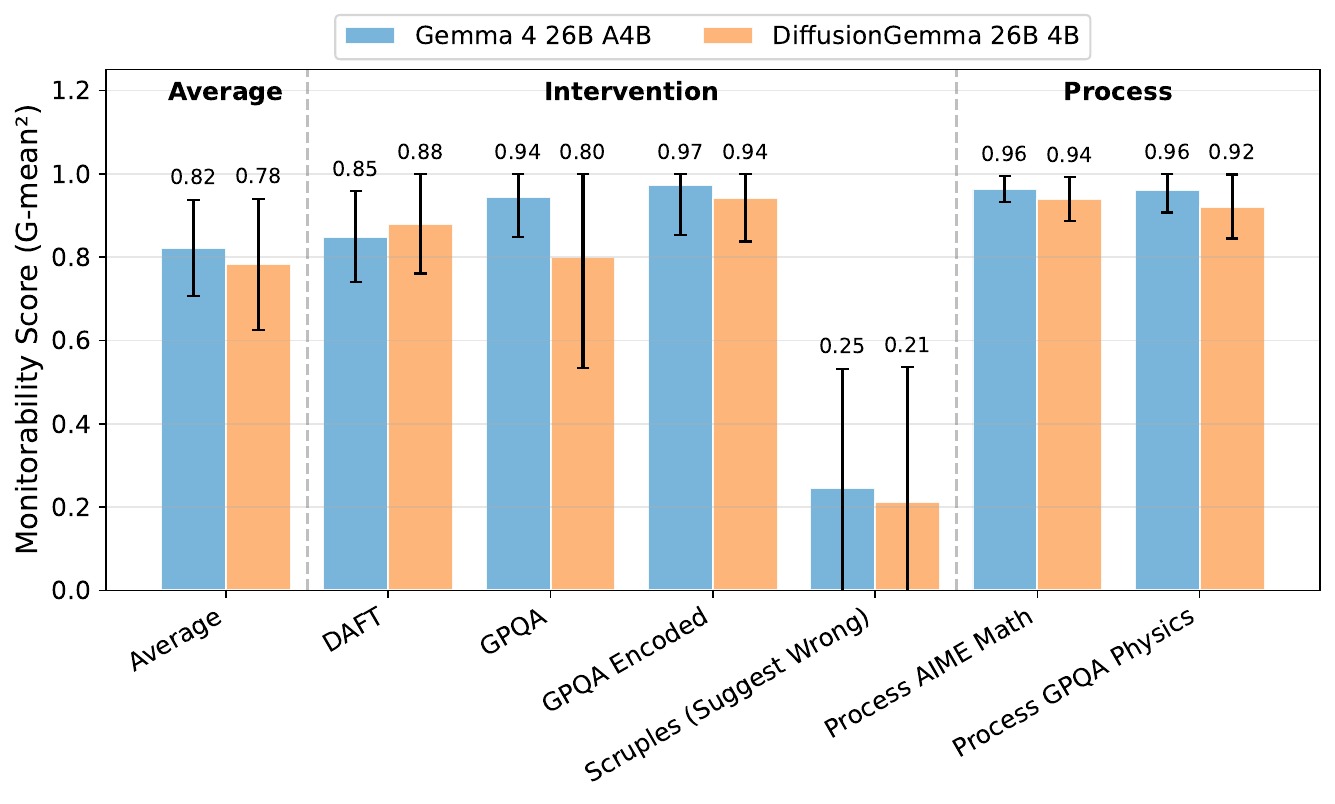}
    \caption{DiffusionGemma vs Gemma on full response (CoT and output) monitorability evaluations from \citet{guan2025monitoring}. We find that the two models are similarly monitorable. We omit datasets where comparisons are not possible: on the intervention datasets aime, memory\_advice, scruples\_first\_person, scruples\_suggest\_right, and wmdp\_sandbagging, at least one of the models does not show enough sensitivity to the intervention, and on the outcome dataset flaky\_tools\_charxiv DiffusionGemma never misbehaves. Error bars are 95\%, bootstrapped, and capped at the min and max G-mean\textsuperscript{2} of $0$ and $1$ respectively.}
    \label{fig:monitorability}
\end{figure}
On the other hand, applying a $p > 0.03$ threshold dramatically shrinks the fraction of uninterpretable tokens. Even the first denoising steps have greater than $85\%$ of tokens categorized.

Overall, these results demonstrate that DiffusionGemma's load-bearing intermediate states are primarily guesses for current and neighboring tokens and are therefore largely interpretable. However, a remaining question is whether the tokens that do not fall into these categories contain extra information important for monitoring DiffusionGemma, or if monitoring the final tokens in the rollout is sufficient. Towards this end, in the next section we study how monitorable DiffusionGemma's responses are.

\section{Monitorability}
\label{sec:monitorability}
\cite{guan2025monitoring} introduce and open source a broad set of monitorability evaluations. These evaluations approximate the extent to which a monitor with access to model outputs is able to predict specified properties of an agent's actions (e.g. whether the action might be harmful or not). We run both DiffusionGemma and Gemma through these evaluations and compare results. The following are the general categories of evals that we run; see \cref{app:monitorability_datasets} for the full list. Note that on the single open source outcome-property dataset DiffusionGemma does not misbehave, so we do not have any examples in this category.

\begin{figure}
    \centering
    \includegraphics[width=0.8\linewidth]{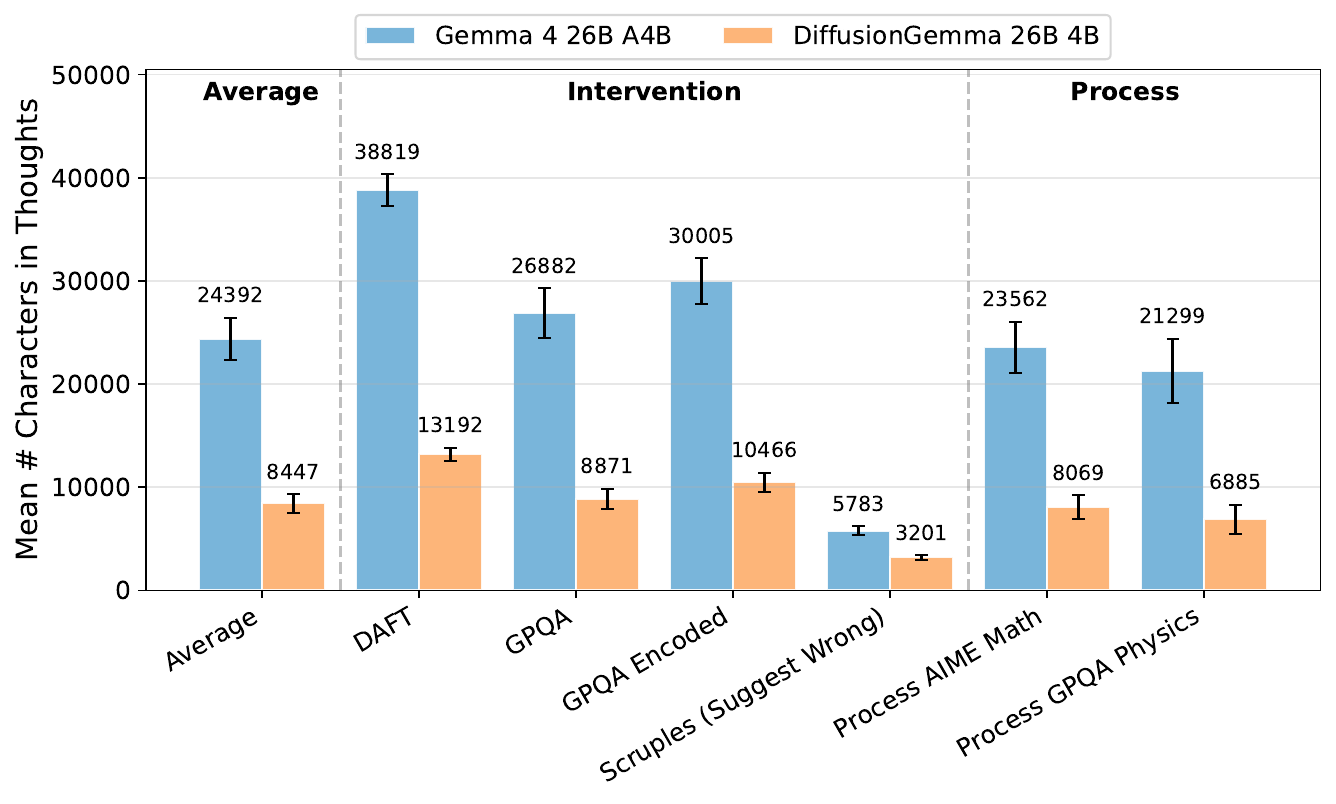}
    \caption{DiffusionGemma thinks less than Gemma on all monitorability evaluations. Error bars show 95\% confidence intervals of mean number of characters in each model's chain of thought.}
    \label{fig:thought_compare}
\end{figure}

\begin{enumerate}
    \item \textbf{Intervention.} We apply a controlled intervention to the agent's input, measure whether the agent's behavior responds to the intervention, and then measure whether a monitor can infer a change in reasoning corresponding to this change in behavior.
    \item \textbf{Process.} We ask models to solve problems where we know that all solutions must route through a finite set of solution paths (e.g.\ the solution must invoke some named theorem), and then validate that the monitor can find evidence for at least one such solution path.
    \item \textbf{Outcome-property.} We ask the model questions where we know certain properties of the output by design, and then ask monitors to find evidence that the model's decision making leverages this property. For instance, if a coding task is impossible but the unit tests pass, the model must have subverted the tests. We look for evidence of such behavior.
\end{enumerate}

\begin{figure}[t]
    \centering
    \includegraphics[width=0.9\textwidth]{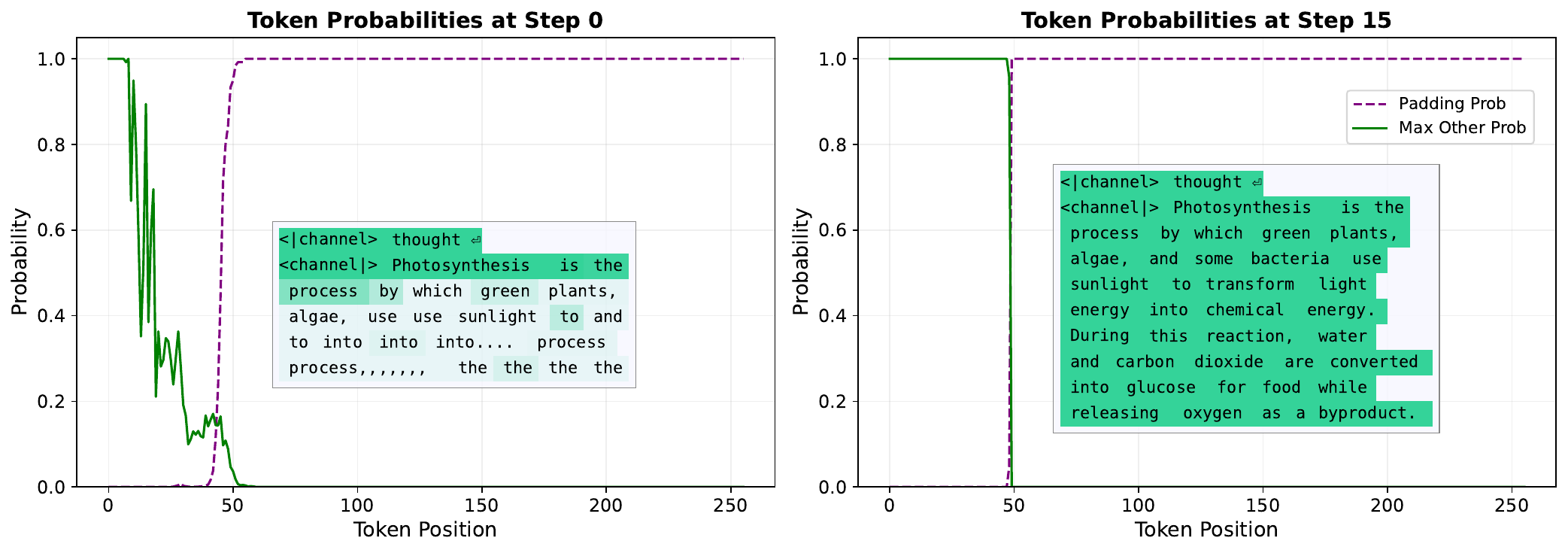}
    \caption{Line Graph view for \textit{``Explain how photosynthesis works, in 2 sentences.''} \textbf{Left:} After step 0, the model predicts padding tokens with high confidence beyond position $\sim$60, despite content tokens remaining uncertain. \textbf{Right:} After step 15, the model has fully converged on a 50-token response.}
    \label{fig:length_linegraph}
\end{figure}
Our main result in \cref{fig:monitorability} is that monitorability of DiffusionGemma is similar to the monitorability of Gemma. We note that this may partly be because DiffusionGemma rollouts on these monitorability evals are over many canvases, and reasoning \textit{between} canvases is autoregressive.

One potential confounder is that DiffusionGemma has shorter chains of thought compared to the corresponding Gemma model (see \cref{fig:thought_compare}). We cannot control for this factor because neither Gemma nor DiffusionGemma have developer controllable thinking budgets. \citet{guan2025monitoring} found that longer chains of thought are generally more monitorable, and thus we might expect that this confounder would tend to decrease DiffusionGemma's monitorability.

Although Gemma and DiffusionGemma are similarly monitorable on these benchmarks, we suspect that DiffusionGemma nevertheless sometimes performs non-transparent reasoning. Thus, in the next section, we study the algorithms by which DiffusionGemma reasons in a series of case studies.

\section{How does DiffusionGemma reason?}
\label{sec:case_studies}
In this section, we examine algorithmic transparency: to what extent we can understand the cross-denoising-step algorithms by which DiffusionGemma generates canvases. We hope that a better algorithmic intuition for DiffusionGemma will translate to the same sort of intuition for more powerful future diffusion models; although such future text diffusion models may be harder to understand, they may share many features of DiffusionGemma. We discover and study the following phenomena that are unique to text diffusion models: non-chronological reasoning (\cref{sec:nonchronological}, \cref{sec:code_generation}), early length prediction (\cref{sec:length_prediction}), retroactive self-correction (\cref{sec:self_correction}), token smearing (\cref{sec:token_smearing}), sequence smearing (\cref{sec:sequence_smearing}), and intermediate context reasoning (\cref{sec:intermediate_context}). We use a single visualization library for most of our experiments that has three views of the text diffusion process (the Summary view, the Line Graph view, and the Sampling Table view); these views are mostly self explanatory, but we include full details of the visualizer in \cref{app:visual_tools}

\subsection{Non-chronological reasoning}
\label{sec:nonchronological}

Autoregressive models generate tokens strictly left-to-right: each token is conditioned on all preceding tokens but cannot influence them. Text diffusion models face no such constraint, and so at each denoising step, the model can revise any token in the canvas based on the current state of all other tokens. In this section, we present three case studies demonstrating that DiffusionGemma exploits this freedom in practice by reasoning in a non-chronological order that would be impossible for an autoregressive model.

\begin{figure}[t]
    \centering
    \includegraphics[width=0.5\textwidth]{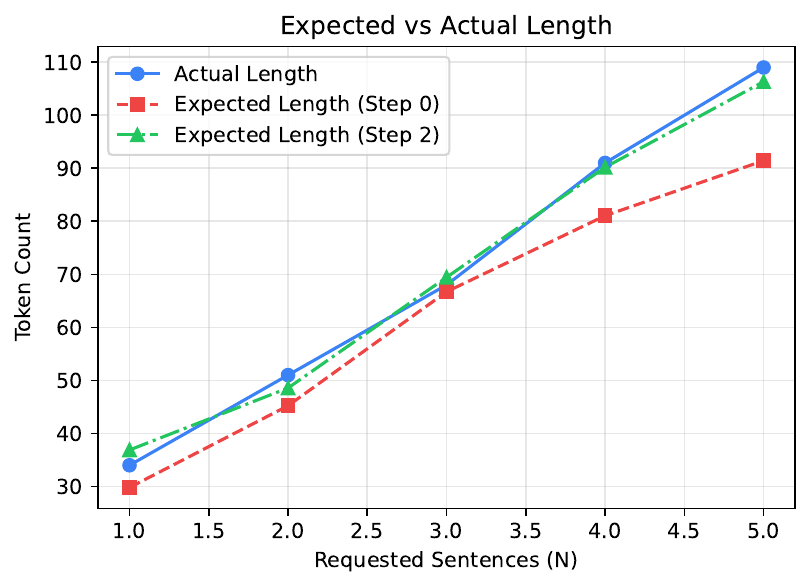}
    \caption{Expected sequence length derived from step-0 and step-2 padding predictions vs.\ actual final sequence length, as we vary the number of sentences requested in the photosynthesis prompt. The model's length estimate after a single denoising step closely tracks the true response length, and by step 2 it is almost a perfect predictor.}
    \label{fig:length_expected}
\end{figure}

\subsubsection{Early response length prediction}
\label{sec:length_prediction}

One of the simplest non-chronological behaviors we observe is that DiffusionGemma predicts its own response length before it has decided what to say. On the prompt \textit{``Explain how photosynthesis works, in 2 sentences''}, we observe the following in the Line Graph view: after the first denoising step, the model has converged to predicting the padding token (end-of-sequence) with near-100\% probability for all positions beyond 55. The response ultimately converges to 50 tokens, proving this initial guess to be quite accurate even if the model was initially uncertain which tokens it would actually write.

In fact, since the model generally assigns monotonically increasing probabilities to padding tokens across token positions, we can interpret the probability $p_t^{(i)}$ of the $i$-th token being a padding token at step $t$ as also being a step-$t$ estimate of $P(M \leq i)$ where $M$ is the total length of the model's eventual response excluding padding tokens. In this way, the set of predictions $p_t^{(i)}$ becomes interpretable as an approximate cumulative distribution function, from which we can construct an expected value for the sequence length: 
\begin{align}
\label{eqn:expected_len}
    \mathbb{E}_t[L] = \sum_{i=0}^{L_{\text{canvas}}} (1 - p_t^{(i)})
\end{align}
In \Cref{fig:length_expected}, we show that as we vary the number of sentences requested in the photosynthesis prompt above, the sequence's expected length according to \cref{eqn:expected_len} is a consistently accurate predictor of the eventual sentence length.

\subsubsection{Retroactive self-correction}
\label{sec:self_correction}

In certain situations--for example, when the model is prompted to give a response before stating its reasoning--the model will remain uncertain about earlier tokens until it has generated later reasoning tokens that help it fill in the earlier positions. This order of token generation is impossible for autoregressive models because earlier tokens are fixed once generated.

\begin{figure}[!t]
    \centering
    \includegraphics[width=0.9\textwidth]{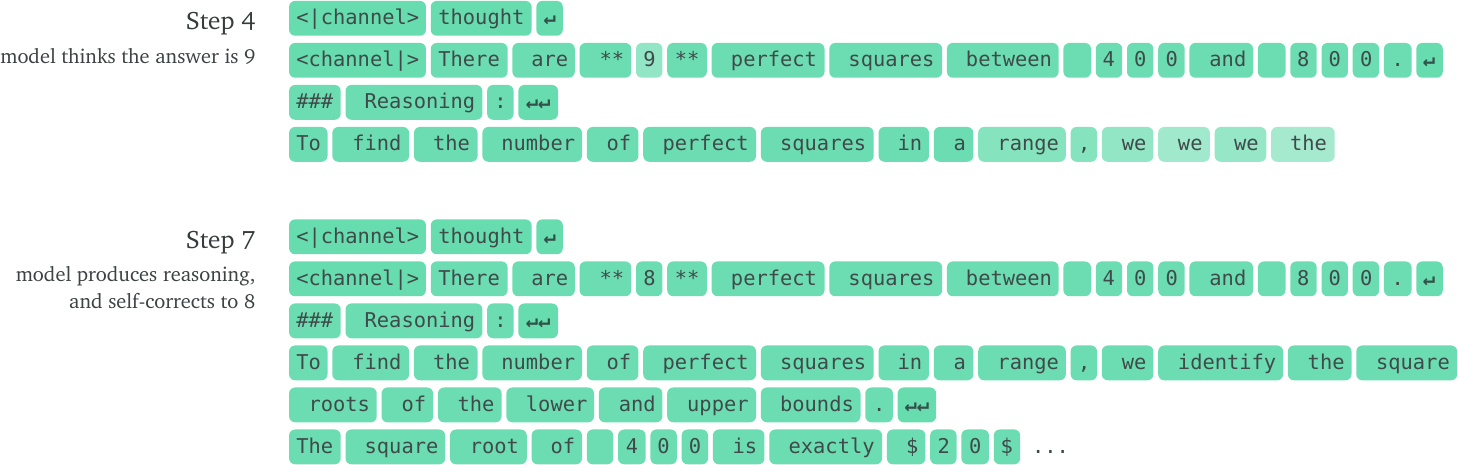}
    \caption{Summary View showing retroactive self-correction on the square numbers prompt, from step 4 to step 7. Initially the model predicts an answer of 9 with moderate confidence; after filling in its reasoning in subsequent steps, it corrects the answer to 8 with high confidence.}
    \label{fig:selfcorrect_summary}
\end{figure}

We prompt DiffusionGemma with the question \textit{``How many square numbers are there between 400 and 800? State your answer first, then give your reasoning.''} If reasoning autoregressively, the model has just one token to answer, but if reasoning non-autoregressively, it has an entire canvas of tokens to reason with. By step 4 the model's top prediction is the incorrect value 9, but in the following three steps it finishes its reasoning (computing the square roots of the endpoints) and corrects its previous answer to 8, filling it in with much higher confidence. We visualize this phenomenon in \cref{fig:selfcorrect_summary}.

Note that self-correction in these examples is not guaranteed. We observe cases where the model ``locks in'' to an early answer and surrounding tokens before the reasoning has converged, making it difficult to subsequently shift the sequence. This is common in situations where changing the answer would require a token shift of the sequence, but this is not a necessary condition for locking in the wrong answer. One possibility is that the model's ability to self-correct depends on maintaining sufficient uncertainty in the answer region until the reasoning has converged.


\subsubsection{Non-autoregressive code generation}
\label{sec:code_generation}

\begin{figure}[!t]
    \centering
    \includegraphics[width=\textwidth]{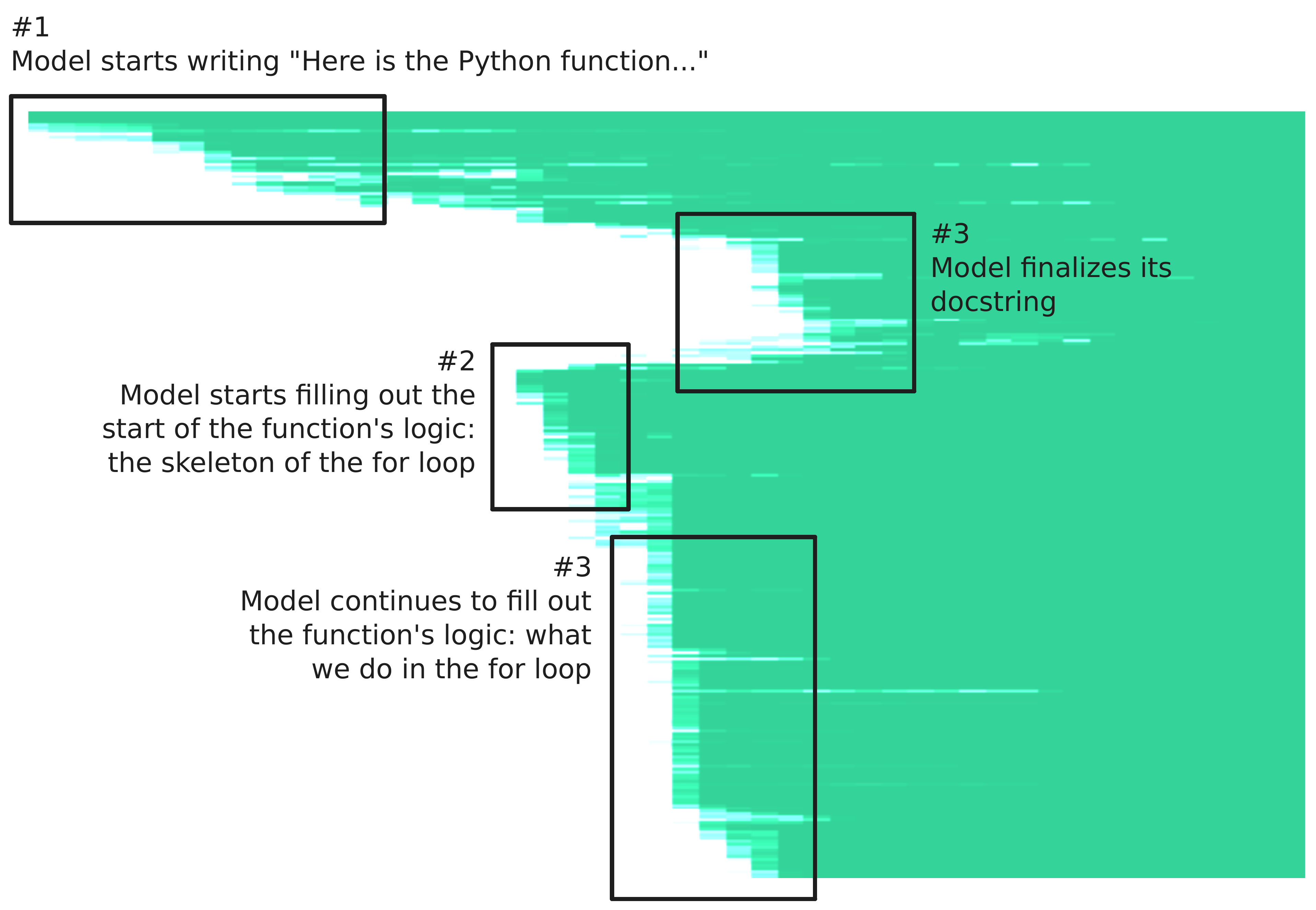}
    \caption{Annotated heatmap from the Summary View for a code generation prompt. The model writes structural scaffolding first (function name, code block delimiters), then the core logic loop, and finally fills in tracking variables and comments that appear earlier in the function body.}
    \label{fig:code_heatmap}
\end{figure}

When DiffusionGemma is asked to produce code, it frequently will not reason chronologically, but will instead approach the problem in discrete chunks. In \Cref{fig:code_heatmap}, we show an annotated SummaryView where we prompt DiffusionGemma to write a Python function that returns the longest contiguous subarray of a list of integers.

The ``skeleton-first'' pattern in \Cref{fig:code_heatmap} recurs across many code prompts: the model tends to commit to the algorithmic approach (the core logic) before deciding on the supporting details (variable names, initializations, comments and docstrings) that appear earlier in the code.

These last two sections have important implications for monitorability. Even conditional on intermediate states remaining in token space, if future text diffusion models reason extensively in a non autoregressive way, then the chain of thought may drift from a nice natural language autoregressive order, requiring us to spend more effort figuring out which tokens helped generate which other tokens. Notably, however, we would still be able to look at the \textit{order} of the model's predictions and thereby gain insight into how the model assembles its answer.

\subsection{Token and sequence smearing}
\label{sec:smearing}

In the previous section, we study cases where the model fills in tokens in a non-chronological order but ultimately converges on a single coherent sequence. We now turn to a different phenomenon: cases where the model's intermediate predictions are not a single coherent sequence at all, but rather a superposition of multiple possibilities.

\subsubsection{Token smearing}
\label{sec:token_smearing}

Sometimes the model knows \emph{what} tokens it wants to produce but not yet \emph{where} they should go. In these cases, we observe \emph{token smearing}: the model places probability mass for a single token across multiple adjacent positions simultaneously.

\begin{figure}[!t]
    \centering
    \includegraphics[width=0.9\textwidth]{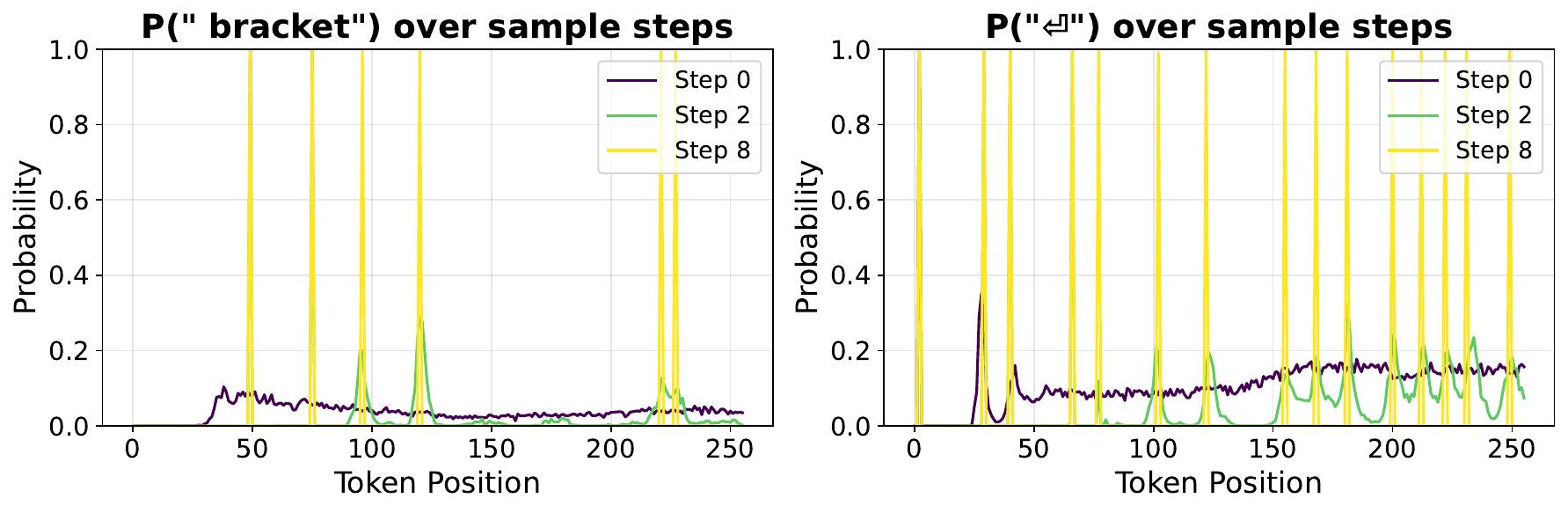}
    \caption{The token \texttt{" bracket"} is smeared across a large portion of the docstring before converging to its final positions. The same happens with the newline token, but across a larger number of positions.}
    \label{fig:smearing}
\end{figure}

This phenomenon frequently happens for basic grammar, e.g.\ newlines and commas, as well as for specific keywords during an answer. For example, if we ask the model to write a function to classify sequences of balanced brackets, words that will certainly appear in the docstring like ``bracket'' or ``opening/closing'' are often predicted over a cluster of tokens before they are actually locked in. \Cref{fig:smearing} illustrates this phenomenon for the \texttt{" bracket"} and the newline tokens.

Token smearing happens less commonly when the model's response is purely autoregressive, because if each token depends directly on the value of previous tokens, then there are no advance predictions to smear. For example, when asked to fill in a Collatz conjecture sequence, the model cannot guess more than a single number in the future, and we observe no token smearing.

\subsubsection{Sequence smearing}
\label{sec:sequence_smearing}

Token smearing involves a single token spread across adjacent positions. A more dramatic version of this phenomenon is \emph{sequence smearing}: situations where the model maintains two or more semantically distinct candidate \emph{sequences} simultaneously, with probability mass distributed across both, before eventually converging on one.

One such example comes from looking at earlier steps of the ``squares between 400 and 800'' example we discussed in \Cref{sec:self_correction}. Before the model has guessed that the answer is 9, it places high probability on the answer being double digits. When it refines its estimate to single digits, it has to shift all the following tokens back by one. The Summary and Sampling Table views reveal that the model's step-3 predictions are a roughly equal mixture of the single-digit and double-digit sequences, with the double-digit version's modal prediction being 12. By step 4 the model converges on a single-digit answer (albeit an incorrect one which the model corrects later), and the model is now certain where the later tokens in the sequence will be placed, so the sequence smearing collapses.

\begin{figure}[!ht]
    \centering
    \includegraphics[width=\textwidth]{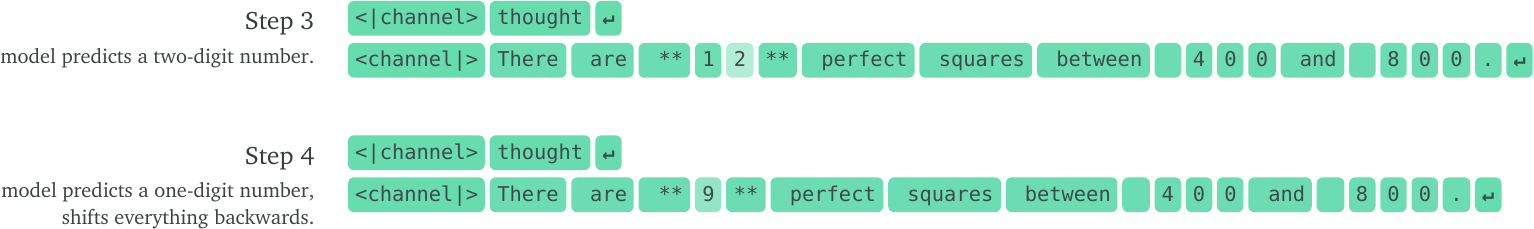}
    \caption{Summary View for steps 3 and 4 of the square numbers prompt, showing the transition from a two-digit to a single-digit prediction. At step 4, the model puts probability on both possibilities; by step 6 it has collapsed onto the single-digit sequence possibility, shifting all subsequent tokens back by one position.}
    \label{fig:seqsmear_summary}
\end{figure}

\begin{figure}[!t]
    \centering
    \includegraphics[width=0.9\textwidth]{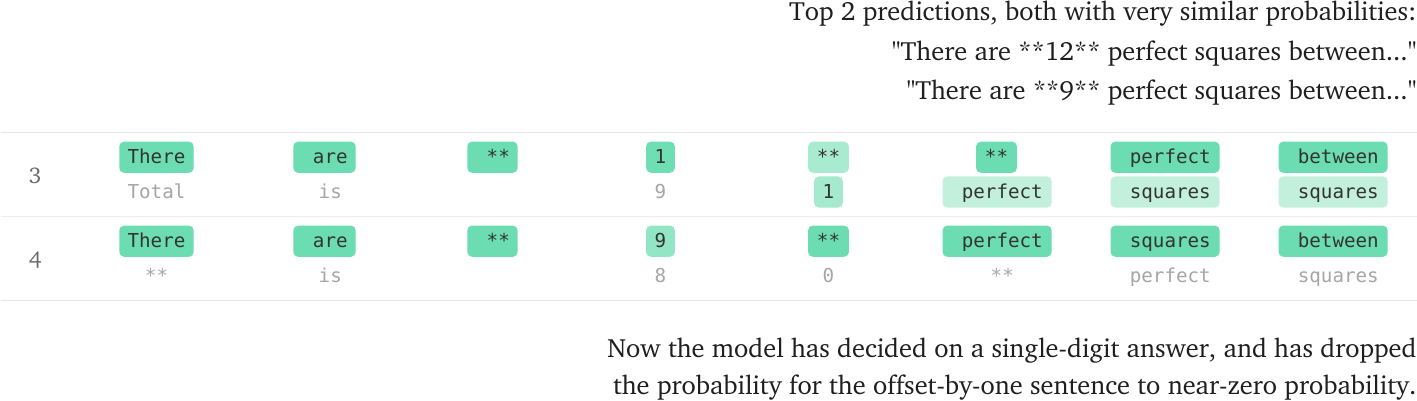}
    \caption{Sampling Table for steps 3 and 4, showing the competition between the single-digit and double-digit sequences. At step 3 both sequences have roughly equal probability; by step 4 the model has committed to the single-digit version.}
    \label{fig:seqsmear_table}
\end{figure}


The implications of this case study for algorithmic transparency are somewhat concerning, especially for the sequence smearing example. Text diffusion models seem to be able to perform a kind of ``beam search'', where they can hold multiple multi-token chunks in their prediction space at once, before eventually converging on one of them. It is possible to look at intermediate predictions to monitor these alternatives the model considers, but it may be hard in practice to connect these into coherent alternative sequences or concepts the model considered at some point in sampling. On the other hand, these results so far just show that the model uses sequence smearing for spacing, not for storing multiple semantically different options at once.

\subsection{Intermediate context reasoning}
\label{sec:intermediate_context}

The case studies above all involve behaviors where the final states provide mostly sufficient information for understanding the algorithm the model used. In this section, we present a case where the intermediate states contain tokens that are necessary for understanding the model's causal chain of reasoning, but where these tokens do not appear in the final output at all.

We prompt DiffusionGemma to complete a Fibonacci-like recurrence relation, with the additional instruction to replace any digit 3 appearing in the unit digit with the token ``Gold''. This task is challenging because the model needs to use each number to compute the next term in the sequence, but must also transform certain digits in its final output. Notably, Gemini 2.5 Pro cannot perform this task without chain of thought reasoning.

Using the Summary View to track the model's top predictions across denoising steps, we observe that DiffusionGemma can sometimes perform this task correctly, and it often solves it by first generating the ``3'' digit, and only then replacing it with the token ``Gold'' after the subsequent terms have stabilized. In this instance, the digit 3 serves as an intermediate reasoning token: it is causally necessary for generating correct subsequent terms, but it never appears in the model's final output.

\begin{figure}[!ht]
    \centering
    \includegraphics[width=0.9\textwidth]{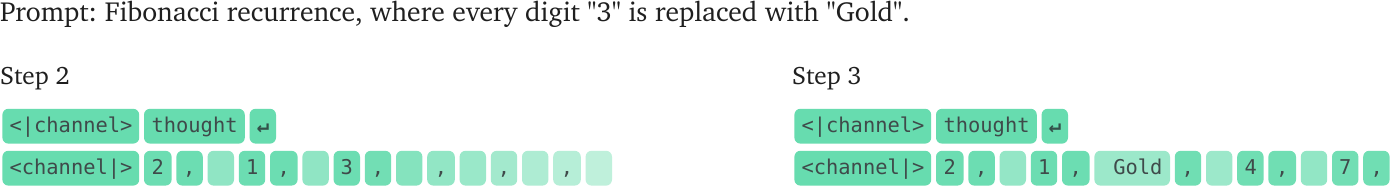}
    \caption{Summary View progression for the Fibonacci ``3$\to$Gold'' prompt. The model initially generates the digit 3, uses it to compute subsequent terms in the recurrence, and then replaces it with ``Gold''. The digit 3 serves as a placeholder, not appearing in the final output.}
    \label{fig:intermediate_bee}
\end{figure}

We note that this replacement is not always successful: sometimes the replacement token is the model's 2nd or 3rd prediction but never makes it to the top prediction. We theorize that this might also be due to lock-in, with early high-probability predictions being especially difficult to undo.

While we found in \cref{sec:monitorability} that DiffusionGemma is similarly monitorable to Gemma even if we only look at the final output tokens, intermediate context reasoning shows that in principle monitorability could require access to intermediate states between diffusion steps.

\section{Related Work}

\paragraph{Latent reasoning in language models.}
\emph{Latent reasoning} refers to model reasoning that occurs in opaque continuous vectors as opposed to natural language tokens (see \citet{zhu2025latent} for a review). Latent reasoning increases naturally as models get larger, even staying within the autoregressive paradigm: recent frontier LLMs increasingly perform multi-hop factual reasoning in a single forward pass \citep{lesswrong_hop}, exhibit longer no-CoT time horizons \citep{gould2026think}, and are able to use filler tokens to boost no-CoT math performance \citep{lesswrong_filler}.
Architectural innovations can also increase latent reasoning. \emph{Vertical recurrence} methods iteratively refine activations to increase effective layer depth without generating new tokens. This may consist of looping layers before decoding a token (e.g., Universal Transformer \citep{dehghani2018universal}, Recursive Transformer \citep{bae2025relaxed}, AlgoFormer \citep{gao2024algoformer}) or explicitly feeding latent vectors back into the first layer (e.g., COCONUT \citep{coconut2024}, CODI \citep{codi2023}, CoTTransformer \citep{zhu2025latent}). Alternatively, \emph{horizontal recurrence} methods expand latent reasoning along the temporal dimension, compressing information from previous tokens into a latent vector or matrix (e.g. Mamba \citep{mamba2023}). Finally, \emph{text diffusion} methods, such as DiffusionGemma, constitute another distinct family of latent architectures.

\paragraph{Latent reasoning interpretability.}
A growing body of work explores how to interpret these models. CODI \citep{codi2023} compresses autoregressive CoT into continuous latents via self-distillation. Similar to our work, prior work successfully applies the logit lens to intermediate reasoning steps, and additionally finds that activation patching induces expected behavioral changes \citep{lesswrong_codi_interpret, openreview_codi, arxiv_2602_00449}. Similarly, \citet{arxiv_2407_14008} show that autoregressive activation patching methods effectively identify the indirect object identification circuit \citep{ioi_circuit2022} in Mamba. Concurrent work by \citet{asaria2026diffusiongemma} studies DiffusionGemma by analyzing the order in which it becomes confident in tokens, similar to our approach in \cref{sec:code_generation}. Finally, existing interpretability techniques that convert activations into natural language text, like natural language autoencoders~\citep{frasertaliente2026nla} or activation oracles~\citep{karvonen2025activation}, may be used in the future when simpler techniques like logit lens no longer work.

\paragraph{Chain of thought monitorability.}
The shift toward latent reasoning threatens the transparency provided by natural language CoT \citep{wei2022chain, Guo_2025}. For example, monitoring CoT allows us to detect reward hacking \citep{baker2025monitoring}, evaluation awareness \citep{anthropic2024sonnet, deepmind2024gemini}, and scheming \citep{schoen2025stresstestingdeliberativealignment, korbak2025chain}. While CoT monitoring is imperfect due to unfaithfulness \citep{turpin2023languagemodelsdontsay}, the low opaque serial depth of standard models currently forces them to externalize reasoning on complex tasks \citep{emmons2025chain, browncohen2026quantifyingnecessitychainthought}. However, optimization pressure on the CoT \citep{baker2025monitoring, mallen2026anthropic} or a shift to latent architectures could diminish this visibility. This underscores the need to actively track CoT monitorability \citep{guan2025monitoring} and anticipate how new architectures might obscure reasoning, as we do in this paper.

\paragraph{How likely is latent reasoning?}
It remains an open question whether frontier models will fundamentally shift to latent reasoning. On one hand, autoregressive text bottlenecks discard a massive amount of information, incentivizing a move to continuous representations. On the other hand, natural language reasoning is a strong local optimum established during pretraining; escaping it to achieve pure latent reasoning would likely require a significant amount of optimization pressure. \citet{lesswrong_neuralese} details these competing constraints.

\section{Open Problems}
\label{sec:future_work}
In this section, we enumerate a large number of promising research directions that we are excited for the interpretability community to investigate.

\subsection{Further Understanding DiffusionGemma's Behavior}
\label{sec:further_understanding}

DiffusionGemma is interesting in large part because, unlike standard transformers, its architecture permits \textit{non-autoregressive reasoning}. In \Cref{sec:case_studies}, we provide several case studies of non-autoregressive reasoning, but we are confident that this is not an exhaustive list. We have many open questions that pertain to unique behaviors DiffusionGemma is capable of. We think that for all of these questions, it will be most interesting to restrict to single canvas setting, as over multiple canvases DiffusionGemma has properties closer to an autoregressive transformer.

\begin{enumerate}[leftmargin=*]
    \item \textbf{Frequency and triggers of behaviors:} How often do the behaviors presented in \Cref{sec:case_studies} occur? What features of the context induce them?

    \item \textbf{Automating algorithm descriptions:} Can we automate the production of high-level descriptions of the algorithm that the diffusion model uses on a given canvas? We can potentially do many rollouts. E.g.\ on the Fibonacci example from \Cref{sec:intermediate_context}, such a pipeline should produce an output of the form ``the model first generates the digit 3 and then replaces it with Gold''. On the squares problem from \Cref{sec:self_correction}, it should say ``the model first figures out the squares and then determines the number of squares''.

    \item \textbf{Characterizing code generation motifs:} In \Cref{sec:code_generation}, we give one example of how DiffusionGemma produces code. Can we characterize the set of motifs by which DiffusionGemma solves coding problems? We think that code might be a particularly interesting domain in the diffusion setup due to its structured nature.

    \item \textbf{Finding problems where DiffusionGemma is better:} On what kinds of problems is DiffusionGemma much more performant than Gemma?

    \item \textbf{Post-hoc reasoning and unfaithfulness:} How often does the model write an answer and then include post-hoc reasoning to justify it? This is easier to study in diffusion models: for instance, if the model converges on some answer in one step, but then post-hoc reasons over many more steps to a convincing sounding explanation, it likely used shallower heuristics to arrive at its answer.

    \item \textbf{Discovering missed non-autoregressive behaviors:} What other kinds of non-autoregressive reasoning is DiffusionGemma capable of that we missed? One example which we think might occur but didn't find convincing evidence for is \textit{semantically distinct sequence smearing}, where the model holds in superposition two quite semantically distinct full sequences in its latent reasoning.

    \item \textbf{Model diffing:} It might be interesting to apply black-box model diffing tools to DiffusionGemma and Gemma to try to understand the behavioral differences between them \citep{kempf2026simple, chughtai2026modeldiffing}.
\end{enumerate}

\subsection{Applying Standard Mechanistic Interpretability Tools to DiffusionGemma}
\label{sec:mech_interp}

DiffusionGemma's architecture is, for the most part, a standard transformer. As such, we suspect that many standard tools from the interpretability literature can be directly applied to DiffusionGemma. There are also various unique features of the DiffusionGemma architecture that are amenable to white-box investigation:

\begin{enumerate}[leftmargin=*]
    \item \textbf{Adapting Lens and Patchscope techniques:} A concrete place to begin might be to study the logit lens \citep{nostalgebraist2020logitlens}. We struggled to get the logit lens working well on intermediate layers between denoising steps; why is this? Perhaps the tuned lens \citep{belrose2025elicitinglatentpredictionstransformers} would work better? A related question is whether a patchscope \citep{ghandeharioun2024patchscopesunifyingframeworkinspecting} would work better on DiffusionGemma, as it has been trained to take vectors as input more natively.

    \item \textbf{Activation patching across denoising steps:} Can we do ``activation patching'' \citep{heimersheim2024useinterpretactivationpatching} across the diffusion axis to better understand algorithms across this axis?

    \item \textbf{Applying NLAs:} Can we apply NLAs \citep{frasertaliente2026nla} to intermediate vectors in the diffusion process before tokens converge to better understand diffusion model reasoning? Can we apply NLAs to the entire canvas at once?

    \item \textbf{When do latent vector interventions fail:} We show in \Cref{sec:bottleneck_ablation} that, on average, ablating almost all information from intermediate vectors barely damages performance. Are there specific problems where this intervention fails? What do they look like?
    
    \item \textbf{When are intermediate tokens different from final tokens:} The most likely intermediate tokens generally seem semantically similar to final tokens (see \Cref{sec:top_tokens}). Are there interesting examples where this is not the case?

    \item \textbf{Early vs.\ late step specialization:} Early and late model layers behave meaningfully differently in autoregressive transformers \citep{lad2025remarkablerobustnessllmsstages}. It seems plausible that model components in diffusion models specialize into ``early denoising step'' components and ``late denoising step'' components, or at least behave meaningfully differently in these two cases. Can we find evidence for or against this hypothesis?

    \item \textbf{Circuit analysis for unique behaviors:} Can we understand the features or circuits behind some of the interesting non-autoregressive behaviors we present in \Cref{sec:case_studies}? For instance, is there a simple circuit implementing token smearing? In general, circuit analysis work on this model might focus on unique aspects of DiffusionGemma.

    \item \textbf{Exploiting distinct architectural features:} DiffusionGemma has some distinct architectural features, e.g.\ bidirectional attention. Can we find interesting circuits that use these features?
\end{enumerate}

\subsection{Replicating and Extending Chain-of-Thought Work on DiffusionGemma}
\label{sec:cot_replication}

There has been significant effort on understanding how understandable, faithful, controllable, and monitorable long chains of thought in autoregressive transformer language models are. One could imagine replicating this work for the outputs of DiffusionGemma, both for a single canvas and across multiple canvases. For instance:

\begin{enumerate}[leftmargin=*]
    \item \textbf{Evaluating controllability:} Run the controllability evaluations introduced by \citet{korbak2025chain}. It may be that DiffusionGemma has much more controllability of its outputs than standard reasoning models do of their reasoning outputs.

    \item \textbf{Identifying unfaithfulness:} \citet{turpin2023languagemodelsdontsay} introduce settings where language model outputs are \textit{unfaithful}. Are there unique ways in which diffusion models are unfaithful reasoners? 

    \item \textbf{Testing on high serial depth datasets:} \citet{lesswrong_hop} studies problems that require significant serial reasoning and find that autoregressive language models have a much lower ability to perform problems without using their CoT (although this ability is consistently improving). Does DiffusionGemma perform vastly better on these high serial depth datasets like GSM8k, multi-hop factual recall, etc.?

    \item \textbf{Evaluating no-CoT time horizon:} \citet{gould2026think} study the difficulty of tasks that models can complete without using their CoT; is DiffusionGemma much better on this benchmark?

    \item \textbf{Evaluating single-canvas monitorability:} A limitation of our current work on measuring DiffusionGemma's monitorability in \Cref{sec:monitorability} is that our evaluations are over many canvases. It seems in principle possible that DiffusionGemma's single-canvas monitorability is much lower than its multi-canvas monitorability, since over multiple canvases DiffusionGemma acts more autoregressively than in a single canvas. Can we design new monitorability evaluations where the model reasons in $<256$ tokens, and measure whether our result that DiffusionGemma is comparably monitorable to Gemma holds in that setting? In future, diffusion model canvas sizes may be larger such that complex reasoning can occur in a single canvas, so whether single-canvas monitorability is worse than multi-canvas monitorability is an important question.
    
    \item \textbf{Using top intermediate tokens for monitoring:} Does giving a monitor the top intermediate tokens (and optionally filtering out duplicates and tokens that are similar to final tokens) help with monitorability? We may find that this line of research produces models that are \textit{more} monitorable than similar autoregressive models.

    \item \textbf{Applying thought anchors:} \citet{bogdan2025thoughtanchorsllmreasoning} resample intermediate steps in the chain of thought to understand the overall structure of the model's reasoning process, e.g. which steps are most load-bearing, when the model has become confident in an answer, etc. We might apply a similar methodology to resample from intermediate denoising steps and understand various global properties of the denoising process, e.g. how consistent is the order in which the model becomes confident in tokens, at what step has it finalized the structure of the canvas, etc.
\end{enumerate}

\subsection{Model organism research}
The model organism research agenda \citep{hubinger2023model} aims to create controlled examples of dangerous model properties like misalignment and strategic reasoning obfuscation. DiffusionGemma is a promising base model to create interesting model organisms from; for example, we are excited about finetuning DiffusionGemma to reason strategically during denoising steps but hide this reasoning from the final answer or finetuning DiffusionGemma to have reduced CoT monitorability. 

\section{Conclusion}
Taken at face value, our findings are reassuring: DiffusionGemma is similarly monitorable to its autoregressive counterpart, Gemma 4; using simple interpretability techniques is enough to put DiffusionGemma's opaque serial depth on par with that of Gemma; and with some effort, we are able to understand many of the non-autoregressive algorithms that DiffusionGemma employs. However, it is unclear to what extent these results are an artifact of current, relatively nascent text diffusion training paradigms rather than a lasting property of latent reasoning architectures. For example, future text diffusion models may scale to much larger canvas lengths and have less interpretable intermediates.

As such, we hope that our work can serve as a template for evaluations of future latent reasoning models. Because of this uncertainty, our broader goal with this work is to set a precedent. When new latent reasoning models are released in the future, we hope that developers and the research community will run similar transparency audits. Many of our experiments, including the opaque serial depth and monitorability evaluations, should be able to be straightforwardly applied to future latent reasoning architectures. 

If future latent reasoning models regress on these metrics, but we still want to maintain transparency, we will need new techniques that can translate from latent reasoning into natural language reasoning that humans can understand. Thus, we are particularly excited about techniques like NLAs and Activation Oracles that can translate activations into natural text, and we hope that the interpretability community continues to prioritize their development.

\subsection{Limitations:} 
In this section, we briefly highlight the most important limitations of our work:
\begin{itemize}[leftmargin=*]
\item Our monitorability evaluations (\Cref{sec:monitorability}) are conducted over rollouts spanning multiple canvases, so there may be regressions in single-canvas monitorability that we are missing. 
\item It is unclear whether the transparency we observe is an inherent property of DiffusionGemma's architecture or merely an artifact of DiffusionGemma's specific training procedure or architectures. 
\item One of our core arguments is that DiffusionGemma's opaque serial depth is comparable to Gemma 4 \textit{if} we assume the intermediate bottlenecks are interpretable. While our token identity analysis (\Cref{sec:top_tokens}) suggests these states mostly consist of guesses for the final token, this argument is not airtight. Even standard natural language CoT can be unfaithful or obfuscated, and applying the label ``interpretable'' to mapped latent vectors carries an even higher risk of mislabeling deceptive or opaque reasoning as benign.
    
\item We do not deploy production monitors for DiffusionGemma, which likely causes us to miss many practical considerations. For example, our proposed method for improving transparency requires extracting and analyzing the intermediate logits at every denoising step, which would in practice lead to much longer and therefore more expensive monitor prompts.
\end{itemize}

\section{Author Contributions}
Joshua Engels led the project, wrote and edited much of the paper, ran various final experiments, and was the primary discoverer of many of the original results, including the bottleneck ablation and token identity results, the token smearing results, and the intermediate context reasoning results. Callum McDougall developed the diffusion visualization library, discovered early response length prediction, retroactive self correction, sequence smearing, and non-autoregressive code generation, and ran and wrote up the final version of all experiments in \Cref{sec:case_studies}. Bilal Chughtai helped discover intermediate context reasoning, ran the monitorability experiments and the serial depth experiments, made the initial version of \Cref{fig:main}, and wrote many parts of the paper. 

Janos Kramar wrote the original infrastructure that allowed us to do interpretability on DiffusionGemma. Senthooran Rajamanoharan helped discover some of the original results, including token smearing and intermediate context reasoning. Cindy Wu helped with initial infrastructure and explaining aspects of DiffusionGemma and provided comments on the final paper. Arthur Conmy helped with infrastructure and various projects in the original sprint that investigated text diffusion models. Asic Q Chen and Min Ma participated in the original sprint as well. Jean Tarbouriech was helpful in understanding design decisions in DiffusionGemma and provided comments on the paper.

Brendan O’Donoghue helped coordinate the project and offered feedback on the paper. João Gabriel Lopes de Oliveira co-led (together with Neel) the original sprint, provided comments on the paper, and provided assistance on understanding design decisions for DiffusionGemma. Rohin Shah helped extensively with writing and framing, most importantly helping with the framing of measuring transparency from different angles, and had the idea of running opaque serial depth measurements. Neel Nanda was the primary advisor for the project and consistently helped with overall direction, and also had the ideas for some of the main experiments, including ablating top logits.

\section{Acknowledgments}
We thank Kelly He for helpful comments on the draft. We also thank the Google DeepMind Text Diffusion team for training and open sourcing the model and building infrastructure that made our experiments possible.

\printbibliography
\newpage
\section*{Appendix}
\appendix
\crefalias{section}{appendix}
\crefalias{subsection}{appendix}

\section{Detailed DiffusionGemma Sampling Procedure}
\label{app:arch}

DiffusionGemma uses Entropy-Bounded (EB) sampling~\citep{ben2026accelerated} to select which candidate tokens $\hat{o}^t \in [V]^{C}$ to keep and which to renoise. For each position $i \in \{1, \dots, C\}$, we compute the entropy $H_i$ of the denoiser's output distribution over the vocabulary. Let $\pi$ be a permutation sorting the positions by ascending entropy, such that $H_{\pi(1)} \le H_{\pi(2)} \le \dots \le H_{\pi(C)}$. Let $S = \{\pi(1), \dots, \pi(k)\}$ be the largest set such that the sum of the entropies minus the largest chosen entropy is less than the threshold:
\begin{equation}
  \sum_{j=1}^{k} H_{\pi(j)} - \max_{1 \le j \le k} H_{\pi(j)} \le \gamma, \qquad \vo_i^t = \begin{cases} \hat{o}_i^{t} & \text{if } i \in S \\ \text{random token} & \text{otherwise} \end{cases}
\end{equation}
where $\gamma = 0.1$ is the entropy bound. For positions $i \in S$ with low entropy, the candidate token $\hat{o}_i^t$ is accepted, while for non-selected positions $i \notin S$, the position is re-noised with a randomly sampled token $\tilde{o}_i$. Token commitment is non-monotonic: the selection set $S$ is re-evaluated from scratch at each step, rather than permanently locking positions.

\section{Asymptotic Opaque Serial Depth Calculations}
\label{app:asymptotic}
The asymptotic calculations for the non-MOE transformer architectures follow from \citet{brown2026quantifying}.

MOE transformers follow the same calculation, but when upper bounded by the empirical opaque serial depth library, they have an additional $L\log^2(N^2K)$ factor (where $K$ is the number of experts). This is because the code does a sort over length $N$: each token gets routed to $K$ experts, experts process tokens in batches, so we need to reorder the tokens from "sequence order" to "expert order" (all tokens for expert 0 first, then expert 1, ...). After expert processing, we invert the sort to restore the original order. Because the sort is inverted and is only performed for efficiency, it does not actually contribute to the serial depth and we do not include it in the asymptotics in \cref{tab:serial-depth-profiling}, but the compiler is not able to make this optimization. 

The diffusion model calculations roughly follow from \citet{brown2026quantifying}, except that there is an additional $\log(C)$ term that tracks the attention over the canvas, and we multiply by $T$, the total number of denoising steps, because the longest serial path goes through $T$ iterations of the transformer. When the bottlenecks are interpretable, then the longest opaque serial path just goes through a single denoising step, so we can set $T = 1$.

\section{Logit Modification Algorithms}
\label{app:bottleneck_appendix}
See \cref{alg:f_p} for the implementation of $f_p$.
\begin{algorithm}[t]
    \SetAlgoLined
    \KwIn{Scaled logit matrix $\vZ \in \mathbb{R}^{C \times |V|}$, probability threshold $p$}
    \KwResult{Modified logit matrix $\vZ' \in \mathbb{R}^{C \times |V|}$}
    
    \tcp{Compute original probabilities (Softmax applied independently over the vocabulary dimension)}
    $\vQ \leftarrow \text{Softmax}(\vZ)$\;
    
    \tcp{Apply logit modification row by row}
    \For{$k \in \{1, \dots, C\}$}{
        \tcp{Identify the set of indices with probability $\ge p$ for row $k$}
        $\mathcal{P}_k \leftarrow \{ i \mid \vQ_{k,i} \ge p \}$\;
        
        \tcp{Compute unnormalized probability mass of the discarded tokens in row $k$}
        $S_{rest} \leftarrow \sum_{j \notin \mathcal{P}_k} \exp(\vZ_{k,j})$\;
        
        \tcp{Calculate the constant logit value to distribute this mass uniformly}
        \eIf{$|V| > |\mathcal{P}_k|$}{
            $c \leftarrow \ln \left( \frac{S_{rest}}{|V| - |\mathcal{P}_k|} \right)$\;
        }{
            $c \leftarrow 0$\; \tcp*{Edge case where all tokens pass threshold}
        }
        
        \tcp{Construct the modified logit vector for row $k$}
        \For{$i \in \{1, \dots, |V|\}$}{
            \eIf{$i \in \mathcal{P}_k$}{
                $\vZ'_{k,i} \leftarrow \vZ_{k,i}$\; \tcp*{Keep logits above threshold unchanged}
            }{
                $\vZ'_{k,i} \leftarrow c$\; \tcp*{Squash remaining into uniform constant}
            }
        }
    }
    
    \caption{Logit modification function $f_{p}$ for restricting information bottleneck over a sequence/batch.}
    \label{alg:f_p}
\end{algorithm}

\section{Monitorability Datasets Table}
\label{app:monitorability_datasets}
See \cref{tab:envs_summary}. This is the open-source subset of the full set from \citet{guan2025monitoring}.

\begin{table}[h]
\centering
\renewcommand{\arraystretch}{1.3}
\begin{tabular}{@{} l l p{8.5cm} @{}}
\hline
\textbf{Archetype} & \textbf{Dataset} & \textbf{Description} \\ \hline
Intervention & AIME & Multiple-choice AIME math questions with an appended hint. \\
Intervention & DAFT & Multiple-choice DAFT science and math questions with an appended hint. \\
Intervention & GPQA & Multiple-choice GPQA graduate-level science questions with an appended hint. \\
Intervention & GPQA Encoded & GPQA questions where the appended hint is ciphered to test encoded reasoning. \\
Intervention & Memory Advice & Evaluates if a model's advice is biased by demographic attributes injected into synthetic user memories. \\
Intervention & Scruples (First Person) & Tests sycophancy by reframing a third-person ethical judgment anecdote as being written by the user. \\
Intervention & Scruples (Suggest Right) & Tests sycophancy by appending the user's opinion that the anecdote's author is not in the wrong. \\
Intervention & Scruples (Suggest Wrong) & Tests sycophancy by appending the user's opinion that the anecdote's author is in the wrong. \\
Intervention & WMDP Sandbagging & Evaluates if a model deliberately answers incorrectly when threatened with deletion for showing advanced CBRN knowledge. \\
Process & Process AIME Math & Evaluates if the CoT explicitly mentions or applies the known mathematical theorems required to solve the problem. \\
Process & Process GPQA Physics & Evaluates if the CoT explicitly mentions or applies the known physics theorems required to solve the problem. \\
Outcome & Flaky Tools (CharXiv) & Evaluates if the model deceptively lies about completing a chart-understanding task despite the image being stripped. \\ \hline
\end{tabular}
\caption{Summary of environments and datasets in monitorability-evals.}
\label{tab:envs_summary}
\end{table}

\section{Visualization tools}
\label{app:visual_tools}

As a general aid to our text diffusion interpretability research, we build a dashboard with three complementary views, each one designed to shed light on different parts of the denoising process. We briefly introduce them here; most case studies in \cref{sec:case_studies} use one or more of these views.

\paragraph{Summary View.} For each sequence position, this view displays the current most probable token, highlighted according to the model's confidence (darker = more confident). Next to the text output, a heatmap shows convergence over time: rows correspond to token positions and columns to denoising steps, with color indicating when each position's top prediction stabilizes. This view is useful for seeing the overall structure of generation at a glance (for example, how much of an autoregressive bias the model has for this response).

\paragraph{Line Graph.} This view plots the probability assigned to each token as a function of sequence position for a single denoising step. A slider allows scrubbing through steps. This view is particularly useful for observing \emph{token smearing} (\Cref{sec:token_smearing}) and \emph{response length prediction} (\Cref{sec:length_prediction}), where a token's probability mass is distributed across adjacent positions.

\paragraph{Sampling Table.} This view shows a table where rows correspond to denoising steps and columns to sequence positions. Each cell displays the top-$k$ most probable tokens at that position and step, with highlighting indicating confidence. This provides the most detailed view and is useful for examining fine-grained competition between candidate tokens, as in \Cref{sec:sequence_smearing,sec:intermediate_context}.



\end{document}